\documentclass[10pt,twocolumn,letterpaper]{article}

\usepackage{wacv}
\usepackage{times}
\usepackage{epsfig}
\usepackage{graphicx}
\usepackage{amsmath}
\usepackage{amssymb}
\usepackage{xcolor}

\usepackage{graphicx}
\usepackage{amsmath}
\usepackage{amssymb}
\usepackage{verbatim}
\usepackage{gensymb}
\usepackage{textcomp}

\usepackage{makecell}
\usepackage{algorithm}
\usepackage[noend]{algpseudocode}
\usepackage{wrapfig,booktabs}
\usepackage{color}
\usepackage[normalem]{ulem}
\usepackage{multirow}
\usepackage{mathtools}
\usepackage{dsfont}
\usepackage{comment}
\usepackage{multirow}
\usepackage{tabu}
\usepackage{array}

\newcolumntype{x}[1]{>{\centering\let\newline\\\arraybackslash\hspace{0pt}}p{#1}}

\newcommand{\textapprox}{{\raise.17ex\hbox{$\scriptstyle\sim$}}}
\DeclareMathOperator{\MAPE}{MAPE}

\def\wacvPaperID{524} 

\wacvfinalcopy 

\ifwacvfinal
\usepackage[breaklinks=true,bookmarks=false]{hyperref}
\else
\usepackage[pagebackref=true,breaklinks=true,colorlinks,bookmarks=false]{hyperref}
\fi

\ifwacvfinal
\fi

\newcommand\blfootnote[1]{%
  \begingroup
  \renewcommand\thefootnote{}\footnote{#1}%
  \addtocounter{footnote}{-1}%
  \endgroup
}

\begin{document}

\title{AI on the Bog: Monitoring and Evaluating Cranberry Crop Risk}

\author{\large Peri Akiva*\textsuperscript{1} \hspace{0.28cm} Benjamin Planche*\textsuperscript{2} \hspace{0.28cm} Aditi Roy\textsuperscript{2} \hspace{0.28cm} Kristin Dana\textsuperscript{1} \hspace{0.28cm} Peter Oudemans\textsuperscript{3} \hspace{0.28cm} Michael Mars\textsuperscript{3}\\
\textsuperscript{1}Department of Computer and Electrical Engineering, Rutgers University\\
\textsuperscript{2}Siemens Corporate Technology\\
\textsuperscript{3}Department of Plant Biology, Rutgers University\\
{\tt\small \{peri.akiva, kristin.dana\}@rutgers.edu \hspace{0.4cm} \{benjamin.planche, aditi.roy\}@siemens.com}\\
{\tt\small \{oudemans, mm2784\}@njaes.rutgers.edu}
}
\maketitle
\thispagestyle{empty} 
\begin{abstract}
Machine vision for precision agriculture has attracted considerable research interest in recent years. The goal of this paper is to develop an end-to-end cranberry health monitoring system to enable and support real time cranberry over-heating assessment to facilitate informed decisions that may sustain the economic viability of the farm. Toward this goal, we propose two main deep learning-based modules for: 1) cranberry fruit segmentation to delineate the exact fruit regions in the cranberry field image that are exposed to sun, 2) prediction of cloud coverage conditions and sun irradiance to estimate the inner temperature of  exposed cranberries. 
We develop drone-based field data and ground-based sky data collection systems to collect video imagery at multiple time points for use in crop health analysis. Extensive evaluation on the data set shows that it is possible to predict exposed fruit’s inner temperature with high accuracy (0.02\% MAPE). The sun irradiance prediction error was found to be 8.41-20.36\% MAPE in the 5-20 minutes time horizon. With 62.54\% mIoU for segmentation and 13.46 MAE for counting accuracies in exposed fruit identification, this system is capable of giving informed feedback to growers to take precautionary action (\eg, irrigation) in identified crop field regions with higher risk of sunburn in the near future. Though this novel system is applied for cranberry health monitoring, it represents a pioneering step forward for efficient farming and is useful in precision agriculture beyond the problem of cranberry overheating.



\end{abstract}
\blfootnote{* Equally contributing co-primary authors}
\vspace{-1.5em}
\vspace{-1.5em}
\section{Introduction}
\label{sec:intro}
\begin{figure}[t!]
    \centering
    \includegraphics[width=0.47\textwidth]{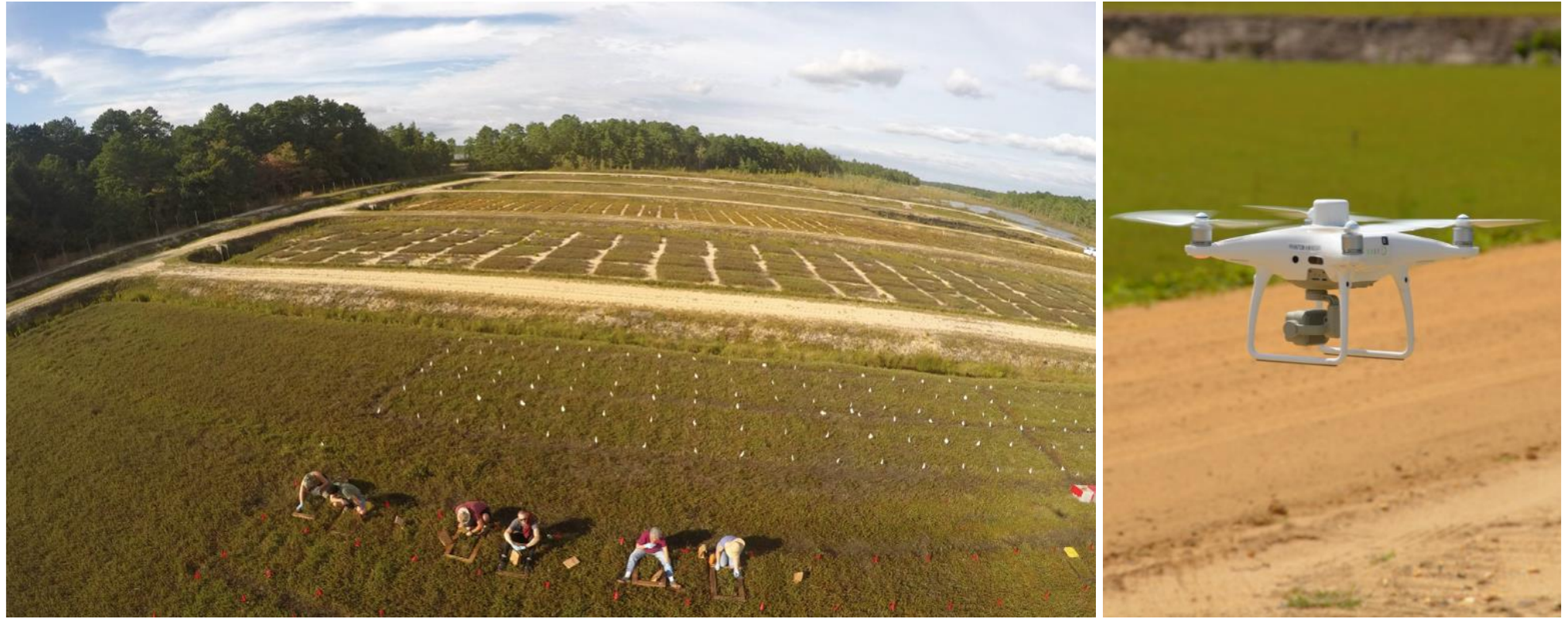}
    \caption{(Left) Aerial view of cranberry field. (Right) Imaging drone. We develop a pipeline for cranberry crop monitoring using drone and sky imagery. Our goal is a computational approach for cranberry growers based on machine learning and computer  vision  to  quantitatively  assess  risk  in  order  to  inform agriculture resource management and enable precision agriculture methods.
    }
    \vspace{-1.5em}
    \label{fig:teaser_v1}
\end{figure}

%
%





\begin{figure}[!tbp]
    \includegraphics[width=0.2\columnwidth]{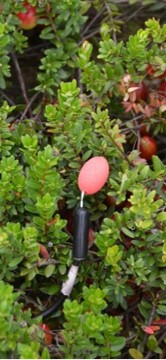}
    \includegraphics[width=0.8\columnwidth]{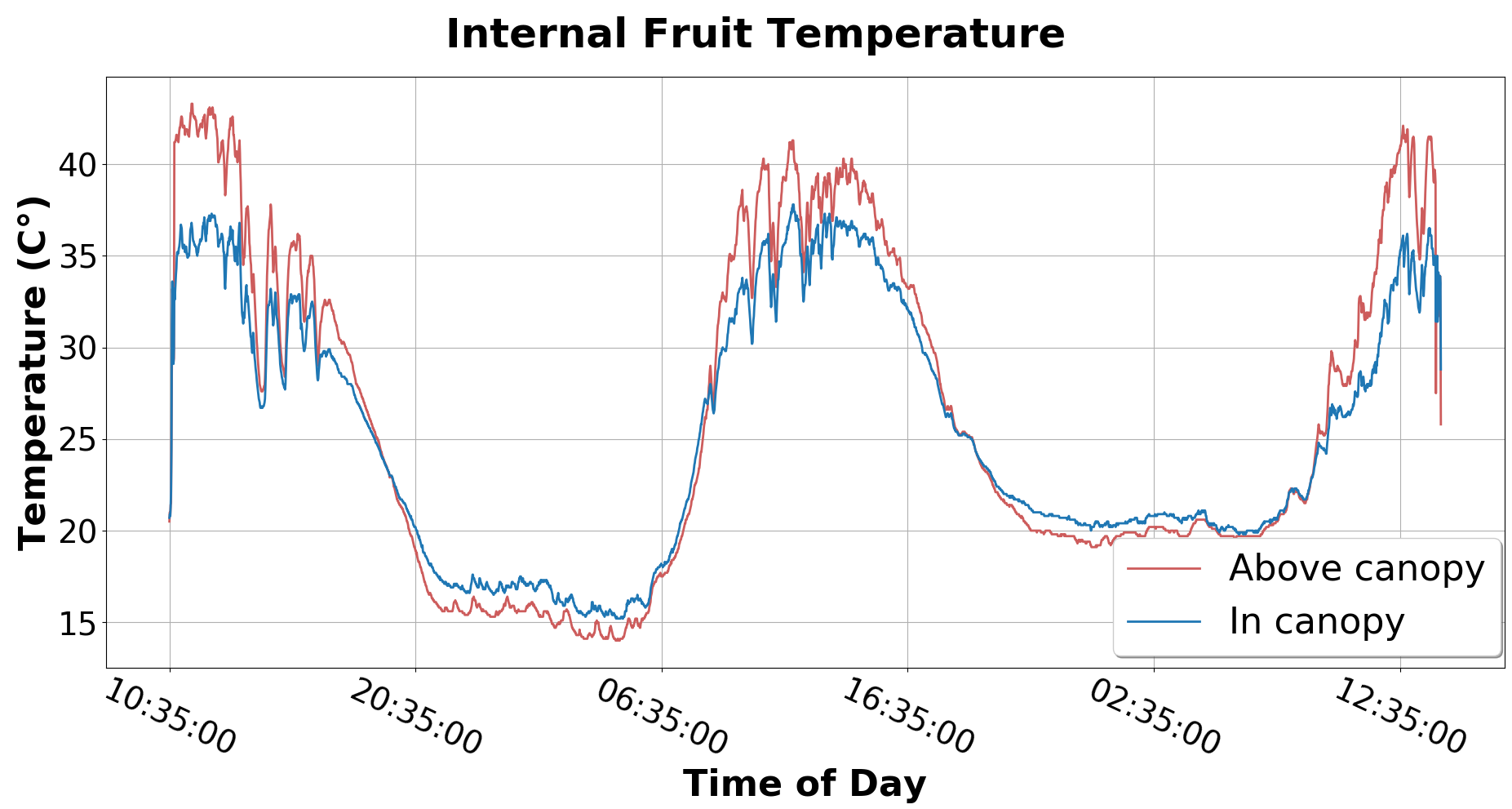}
  \caption{(Left) Synthetic cranberry temperature probe for measuring internal temperatures. (Right) Temperatures measured over time in and above cranberry canopy.}
  \vspace{-1.5em}
\label{fig:teaser}
\end{figure}

%
Precision agriculture aims to improve farming methods and resource management decision making through computational approaches. The long term potential of precision agriculture is in optimizing  yield by real-time monitoring  to precisely tune irrigation, fertilization, and other crop management.  In this work, we focus on the issue of cranberry crop risk assessment to facilitate decisions on irrigation and other crop management methods. 
Sunburn, caused by direct solar radiation, is evidenced by irreversible damage to the tissues of
the cranberry fruit \cite{kerry2017investigating}.
Losses to fruit in the upper part of the cranberry canopy has become
increasingly common over the past 8 years, and in some years, these losses can represent upwards of 30\% of the entire crop in a particular field \cite{grower2020}.
%
Croft \etal \cite{croft1995field} reported that cranberry scald or sunburn occurs infrequently and is limited to areas in New Jersey. However, growing
conditions have changed since that study was conducted. Most importantly, it is possible that the
exposure of cranberry fruit to direct sunlight has increased as cranberry yields have increased
and new, high yielding, large fruited cultivars are being planted.
Preliminary research suggests that the overheating of cranberry fruit is driven by solar radiation
and exposed fruits may experience internal temperatures 25\textdegree C above the canopy temperature \cite{croft1995field} (see Figure \ref{fig:teaser}). Exposed fruits initially soften and then begin to degrade, with fungal fruit rot ultimately developing within.
%
%
%
%

%
Linking sky conditions to berry internal temperature enables a risk assessment since the crop can survive high temperatures only for short time periods before failure.  
The need for quick and accurate feedback is often critical to avoid loss of crop yield \cite{roper2006physiology,kerry2017investigating,pelletier2016reducing}, which may reach up to 100\% loss with sudden environmental condition changes \cite{pelletier2016reducing}.
%
%
This work shows that a combination of computer vision methods may provide an automated system to estimate yield loss and manage irrigation resources. Current yield estimation methods use previous years data or manual counting, while irrigation management relies on weather reports which are infrequent, insufficient and often inaccurate. 

In this work, we develop a three part computer-vision based approach to 1)  automatically segment and count exposed cranberries from down-facing drone imagery (see Figure~\ref{fig:teaser_v1}) to provide near real-time estimates of crops with potential high risk of sun exposure and over-heating; 2) analyze up-facing sky imagery to determine cloud cover and cloud motion for future irradiance prediction and 3) predict internal berry temperature to assess crop risk of over-heating. Segmenting and counting the cranberries enables estimation of the monetary crop value at a given time. By segmenting and counting visible cranberries, the method essentially detects cranberries with direct exposure to sunlight. The counts are then used to obtain density maps of exposed cranberries for an entire field, which may indicate to specific high risk regions that need to be attended during sudden environmental change. By predicting future irradiance with a time horizon of 15-20 minutes through sky monitoring, our method provides farmers a window of opportunity to mitigate over-heating events that may be irreversibly damaging to the fruit. To the best of our knowledge, this is the first approach linking sky observations with field-based fruit condition for its health monitoring.
%
%


\vspace{-0.5em}

\section{Related Work}
\vspace{-0.5em}
{\bf Precision agriculture} using computer vision is growing rapidly in recent years.  
Drones play a significant role in providing an easy framework for efficiently imaging large-scale crop regions and current literature addresses general usage of drones in crop health monitoring \cite{maes2019perspectives, mogili2018review}, drone models on the market \cite{puri2017agriculture}, photogrammetric structure-from-motion 
point clouds of crops from drone imagery \cite{hunt2018good}, automated drone navigation using vision \cite{alsalam2017autonomous}. 
Full 3D environment mapping methods for agriculture, including 3D reconstruction of entire fields \cite{dong20174d,potena2018collaborative,potena2019agricolmap},  often employ a collaboration of aerial and ground vehicles. While global view of fields have utility in long time-frame growth assessment, high resolution imagery to count individual cranberries can be prohibitively data-intensive over large regions.  Toward the goal of real-time crop-monitoring,  we adopt the efficient approach of stratified random sampling for representing large scale regions as in  \cite{pozdnyakova2005spatial}. 

For vision-based analysis of crop health, concentrating on a specific crop  such as coffee \cite{rodriguez2020computer}, olive \cite{gatica2013olive}, grapes \cite{nuske2011yield} and grains \cite{patricio2018computer} has  practical advantages since algorithms can be tuned to expected appearance. This work for cranberry crops  develops a state-of-the-art unique pipeline to process cranberry crop imagery,  sky images and weather data to quantify both the amount of exposed fruit and the level of risk at a given point in time.  While we focus our results on cranberries, we expect that the overall framework is relevant to multiple types of crops using transfer learning, where deep learning network parameters can be fine-tuned for use in other agriculture imagery.
  
%
%
%
\begin{figure*}
    \centering
    \includegraphics[width=\textwidth]{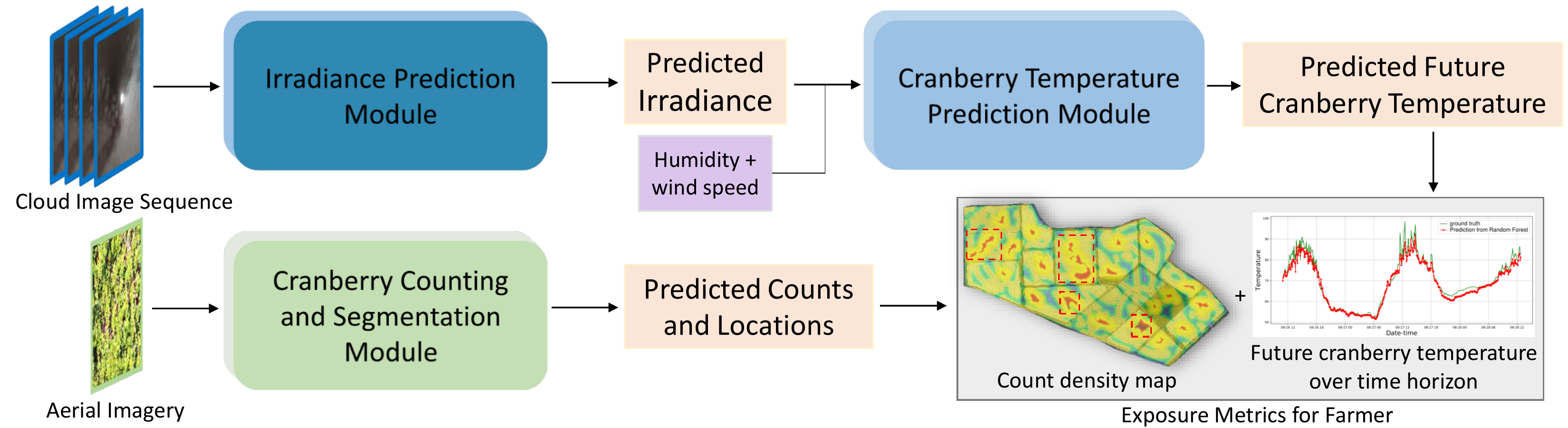}
    \caption{Pipeline overview. Cloud image sequence, humidity, and wind speed are used to predict future berry temperature over a time horizon to determine high risk time periods. Aerial imagery of cranberry crops are used to obtain count density maps to determine high risk regions. Exposure metrics (number of exposed cranberries with high berry internal temperature) are made available to the farmer in order to make resource decision such as crop irrigation. Red dashed boxes indicate high risk regions. Best viewed in color and zoomed.}
    \vspace{-1.3em}
    \label{fig:pipeline_2}
\end{figure*} 

\noindent
\textbf{Deep learning Object Counting} is fundamental for quantifying cranberry exposure, and current
 methods mainly divide into two approaches: counting by regression, and counting by detection. Regression approaches \cite{lempitsky2010learning, chattopadhyay2017counting} tend to be faster as they often explicitly learn count from images by encoding global features
similar to classification methods \cite{simonyan2014very, krizhevsky2012imagenet}. \cite{chattopadhyay2017counting} shows that while counting by regression is faster and accurate in small count scenes, performance degrades with larger count scenes. On the other hand, counting by detection implicitly learns object counts by first detecting every object and its class in the image, and then summing occurrences per class. This implies that perfect detection results in perfect count. While best performing detection methods such as R-CNN and YOLO \cite{girshick2014rich,girshick2015fast,redmon2016you} use expensive \cite{bearman2016s} bounding box annotation, recent approaches relying on weak supervision have shown comparable results. LC-FCN (Localization based Counting Network) \cite{laradji2018blobs} leverages center points annotations to detect small regions used for counting, replacing full object detection with small blobs for each object in the scene. \cite{ribera2019locating} also uses point supervision with a novel Weighted Hausdorff Distance loss to count through localization, relying on an encoder-decoder network for point localization prediction, which is concatenated with the network latent features and fed to a regressor for count prediction. Triple-S Network \cite{akiva2020finding} builds on both \cite{laradji2018blobs,ribera2019locating} by simultaneously predicting count and segmentation using a shape loss prior known about the objects. This work builds on Triple-S Network \cite{akiva2020finding} and \cite{ribera2019locating} to generate density maps of cranberry fields useful for risk assessments and resources management. 
%

\noindent
\textbf{Cloud Monitoring} has been done extensively in the research community over the past few decades for a wide variety of applications, such as nowcasting, to deliver accurate weather forecasts \cite{papinUnsupervisedSegmentationLow2002}, rainfall and satellite precipitation estimates \cite{mahrooghyUseClusterEnsemble2012}, in the study of contrails \cite{weissAutomaticContrailDetection1998}, and various other day-to-day meteorological applications \cite{christodoulouMultifeatureTextureAnalysis2003}. To monitor cloud dynamics in a local area and within a short time horizon of 20-30 minutes, a fish-eye camera-based system is the best choice as it provides high temporal and spatial resolution hemispherical information of the cloud cover by taking the sky pictures continuously every few seconds. From the time series of the images, it is possible to obtain a good estimate of the cloud trajectory and thus predict when and how much the sunlight would be occluded \cite{sunShortTermCloud2014,changShortTermCloud2019}. To estimate cloud trajectory, we need to perform cloud segmentation and motion estimation, which are known to be closely related \cite{changShortTermCloud2019,chengSegFlowJointLearning2017,qinJointLearningMotion2018a}.  
Most existing cloud segmentation techniques are based on color features \cite{heinleAutomaticCloudClassification2010,devColorbasedSegmentationSky2016} and do not emphasize the generalization to the 
stochastic cloud appearance. Recent deep-learning-based popular segmentation algorithms (\eg, U-Net++ \cite{zhou2018unet++}, SegNet \cite{badrinarayanan2017segnet}, \etc) are heavily dependent on finely labeled training data and seldom perform motion estimation jointly. To leverage temporal continuity of image sequences, few approaches \cite{badrinarayanan2017segnet,chengSegFlowJointLearning2017,qinJointLearningMotion2018a,zhuImprovingSemanticSegmentation2019} consider optical flow prediction as an additional task apart from segmentation to make the network robust by synergy. Approaches that exploit the synergy between these two tasks have shown superior results on both tasks. Motivated by this observation, we employ a weakly supervised model for joint estimation of cloud motion and segmentation from sky image sequences.

\vspace{-0.1in}
\section{Method}
\subsection{Problem Formulation and Pipeline}
The application pipeline illustrated in figure \ref{fig:pipeline_2} follows two branches: the berry temperature prediction branch, and the cranberry segmentation and counting branch.  The input to the berry temperature prediction branch are sequences of up-facing (sky) images taken near the cranberry bog to predict future solar irradiance over that region. 
%
Irradiance information is paired with current humidity and wind speed measurements and serves as input into the berry temperature prediction module, which provides predicted future internal berry temperature. As discussed in Section~\ref{sec:intro}, internal berry temperature is a key element in determining berry spoilage, as they become irreversibly damaged beyond certain temperatures. This link between spoilage and temperature makes predicting future internal berry temperature highly valuable in preventing crop loss. Furthermore, cranberry counts provide a quantitative measure of the economic impact of the potential loss. The cranberry segmentation and counting branch also provides locations in which exposed cranberries are at present. Together, the two branches provide information to cranberry growers about high risk times and regions. This information is critical for resource decisions such as whether the cost of additional irrigation is justified by the value of the at-risk crop. 

\begin{figure*}[t!]
    \centering
    \includegraphics[width=\textwidth]{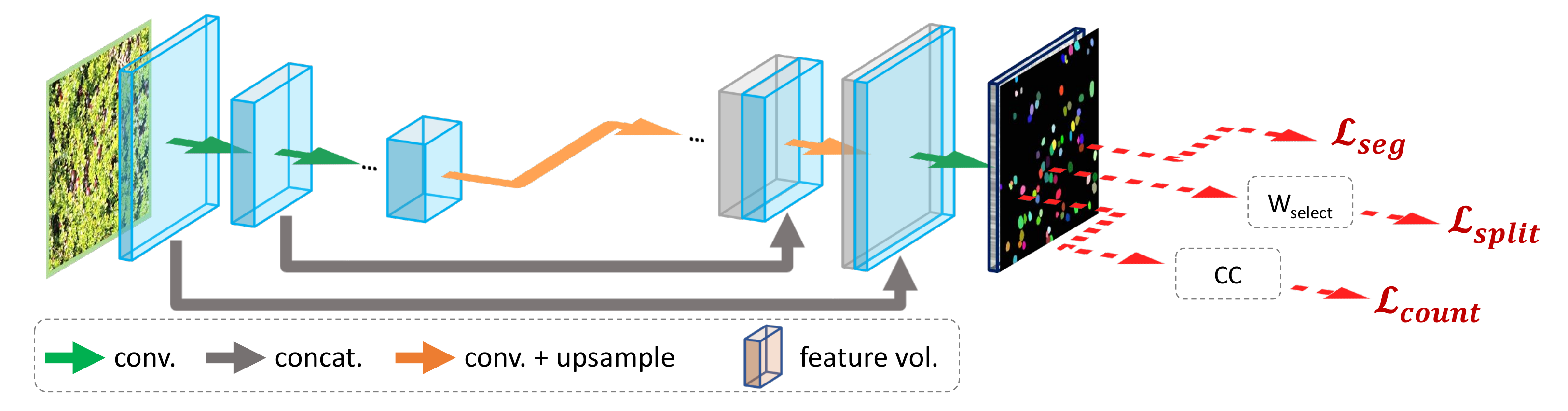}
    \caption{Triple-S Network architecture used in this application. Aerial image is input to a U-Net style network with output guided by segmentation loss, $\mathcal{L}_{seg}$, split loss, $\mathcal{L}_{split}$, and count loss, $\mathcal{L}_{count}$. $W_{select}$ is the selective watershed algorithm introduced in \cite{akiva2020finding}, and $CC$ is the connected components algorithm \cite{dillencourt1992general}. 
    }
    \vspace{-1.8em}
    \label{fig:overall_system}
\end{figure*}
\vspace{-0.4em}
\subsection{Cranberry Counting and Segmentation}
This part of our application is built upon Triple-S Network's \cite{akiva2020finding} count loss ablation study. Let $X$ be an image with corresponding points $P$, containing cranberry (positive) and non-cranberry (negative) points noted as $p_p$ and $p_n$ respectively. The method feeds input $X$ to an encoder-decoder architecture to obtain output $\Tilde{Y}$ which is fed to three branches aiming to optimize segmentation, split, and count losses. The segmentation loss encourages the model towards accurate blob localization using the points as ground truth. This loss is defined as the cross entropy loss, where the target, $Y$, is a pixel-wise mask obtained by labeling the positive and negative points. 

The split loss aims to discourage overlapping instances in segmentation prediction, and define appropriate regions for predicted blobs. The split loss branch uses the selective watershed algorithm, $W_{\textit{select}}$ \cite{akiva2020finding}, to obtain the set of positive and negative regions representing appropriate regions for cranberry and non-cranberry predictions. The loss function is constructed as the cross entropy loss with prediction $W_{\textit{select}}(\Tilde{Y},Y)$.

Lastly, the count loss branch aims to directly optimize training towards the correct number of instances in the input image. This branch is closely related to detection based counting but does not include a detection algorithm. Instead, it defines instances using the network's segmentation prediction, $Y$, to obtain separable blobs defined as instances. To that end, the connected components algorithm \cite{dillencourt1992general}, noted as $CC$,
~is used on output $Y$, to obtain a set of separable blobs $B$. The predicted count $\Tilde{C}$ is defined as the number of separable blobs $B$ in $Y$, otherwise noted as the cardinality of $B$, $\Tilde{C} = |B|$. The count loss is constructed as a Huber Loss \cite{huber1992robust} with ground truth $|p_p|$ and prediction $\Tilde{C}$.

\subsection{Irradiance Prediction}
The goal of the second part of our system is to help predict local weather conditions that would affect the berry internal temperature, \eg, solar irradiance. In this subsection, we therefore justify the importance and challenges of forecasting solar irradiance based on predicted cloud coverage and movement from local sky images.
\vspace{-1.2em}
\paragraph{Cloud Analysis from Sky Images.}
To predict current and near-future sun occlusion, cloud presence and motion should both be estimated from the sky images. We model the first task as a binary pixel-wise classiﬁcation (identifying cloud and non-cloud pixels), and the second task as an optical flow (\ie, 2D pixel motion map) regression. We extend a deep learning method for joint estimation of motion and segmentation from image sequences \cite{qinJointLearningMotion2018a}. 

\begin{figure*}[t!]
    \centering
    \includegraphics[width=\textwidth]{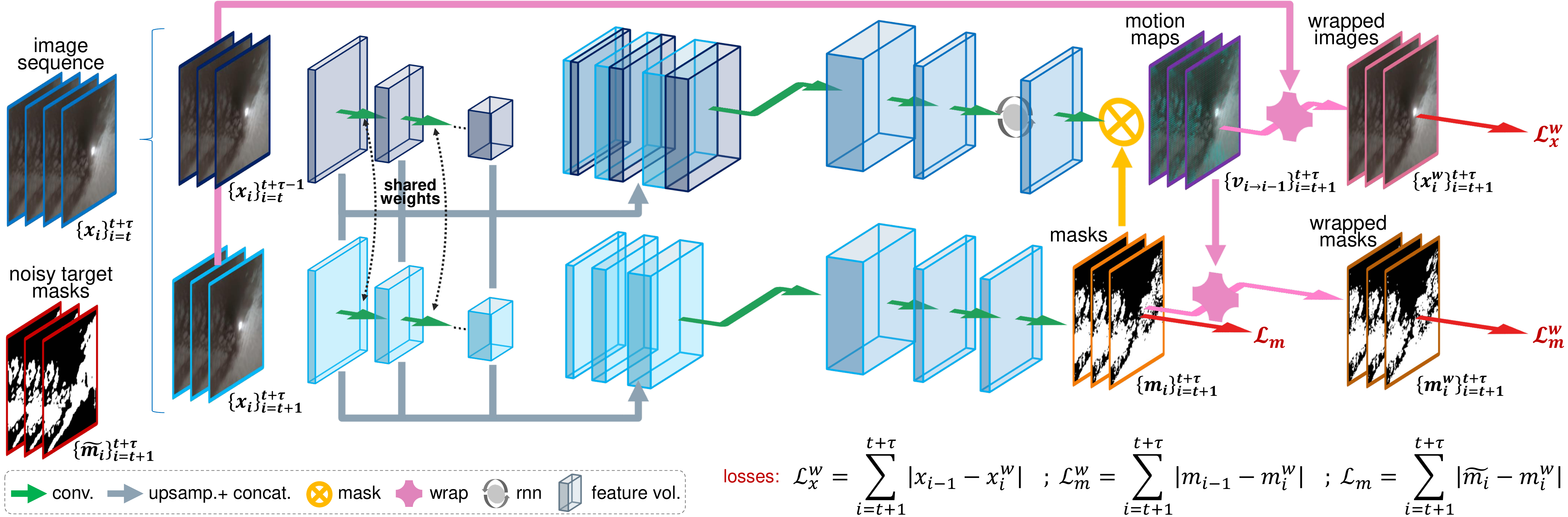}
    \caption{Training of baseline method for joint semantic segmentation and optical flow regression \cite{qinJointLearningMotion2018a} with additional self-supervision.}
    \vspace{-1.5em}
    \label{fig:segframework}
\end{figure*}

As shown in Figure~\ref{fig:segframework}, the baseline network \cite{qinJointLearningMotion2018a} consists of two branches: a cloud motion estimation branch that is built on an unsupervised Siamese style recurrent spatial transformer network, and a fully-convolutional cloud segmentation branch. 
To leverage the synergy between their tasks, the two branches share a joint multi-scale feature encoder \cite{planche2019seeing}. 
This encoder takes as input consecutive sky images $\{x_i \in \mathbb{R}^{H \times W \times 3}\}_{i=t}^{t+\tau}$ and samples them into one sequence of \textit{new} frames $\{x_i\}_{i=t+1}^{t+\tau}$ and one sequence of their \textit{preceding} frames $\{x_i\}_{i=t}^{t+\tau-1}$. 
Features from each consecutive pair $(x_i, x_{i-1})$ (\textit{new} and \textit{preceding}) are concatenated within the motion estimation branch, which then decodes the feature volume into the reverse optical flow $v_{i\to i-1}$ mapping $x_i$ to $x_{i-1}$. Note that a recurrent neural layer is added to the module to refine the flow based on previous predictions. In parallel, the semantic segmentation estimates from the encoded representation of $x_i$ its cloud pixel map $m_i$.

Additionally, we utilize the domain knowledge that, overall, only clouds should be moving from one frame to another in sky images, to increase the synergy between the two network branches. We multiply the predicted per-pixel optical flow values with the corresponding \textit{softmax} cloud probabilities, virtually masking the optical flow for non-cloud pixels.

Since ground-truth semantic masks for cloud segmentation are tedious if not impossible to collect (due to fuzzy cloud contours, irregular cloud density, \etc), we adopt various schemes to weakly supervise the pipeline training.
The joint encoder is learned by optimizing both branches simultaneously. This enables the weakly-supervised segmentation by taking advantage of features that are unsupervisedly learned in the motion estimation branch from a large amount of un-annotated data.

The overall motion estimation branch is trained in a self-supervised manner. For each pair of training images, the next frame $x_i$ is warped by the predicted flow $v_{i\to i-1}$, and the resulting image $x_i^w$ is compared to the preceding target frame $x_{i-1}$ \cf loss $\mathcal{L}_x^w$ in Figure \ref{fig:segframework}. We also augment the baseline solution with an additional loss $\mathcal{L}_x^{w_{rev}} = \sum_{i=t+1}^{t+\tau}{|x_i - x_{i-1}^{w_{rev}}|}$, with $x_{i-1}^{w_{rev}}$ obtained by warping $x_{i-1}$ with the forward flow $v_{i-1\to i}$ predicted by switching the feature concatenation order in the motion estimation branch. This loss ensures backward/forward consistency of predicted optical flows. 
The segmentation branch can be trained via weak supervision (\cf loss $\mathcal{L}_m^w$), as well as direct supervision if target semantic masks $\{\widetilde{m_i}\}_{i=t}^{t+\tau}$ are available (\cf loss $\mathcal{L}_m$). Enforcing temporal consistency and compensating for scarce/noisy target labels, $\mathcal{L}_m^w$ compares $m_i^w$, obtained by warping predicted masks $m_i$ using $v_{i\to i-1}$, to $m_{i-1}$.


\vspace{-1em}
\paragraph{Solar Irradiance Prediction.}
Using the cloud-related predictions, we infer the sun occlusion, and from there the irradiance, for a time interval of 1-20 minutes. To that end, we define a  prediction zone in the sky images that contains the clouds, if any, that will possibly move in to occlude the sun in the considered time horizon. Like \cite{sun2015sun}, we model this region of interest as a 2D band starting at the depicted sun position and extending in the opposite direction of the approximate global cloud motion, computed over the predicted motion maps (see purple rectangles in Figures~\ref{fig:seg_result2}-\ref{fig:ir_result_combined}); and we then apply forward extrapolation in time to estimate the occlusion probability.

The sun position $s_i$ in 2D image space is obtained by projecting the current sun position in 3D sky coordinates obtained from “The Astronomical Almanac” \cite{us1985astronomical}, using the camera's extrinsic and intrinsic parameters.
The global cloud motion vector $\overrightarrow{V} \in \mathbb{R}^2$ has the advantage to average out estimation errors over individual pixels and cloud deformation (which is not true motion), thus achieving a more stable estimate. 
We compute $\overrightarrow{V_i} = 
\sum_{x,y=1}^{H, W}{\gamma_i(x,y) \cdot v_{i\to i+1}(x,y)}$
where $\gamma_i = h(m_i) \cdot f(\|d_i\|_2) \cdot g(v_{i\to i+1}, d_i)$, with $d_i(x,y)= (s_{i,x} - x, s_{i,y} - y)$ and $h$, $f$, and $g$ functions of choices (\eg, exponential or polynomial).
Finally, we determine the clear sky irradiation curve from past observations and use it as the base for computing the irradiance from cloud probability map in the prediction zone. 
\vspace{-0.4em}
\subsection{Cranberry Internal Temperature Prediction}
\vspace{-0.4em}
We  explore a feature selection and regression framework to infer the berry internal temperature from predicted and measured data samples.
The main drivers of berry internal temperature are environmental condition such as solar irradiance, humidity, wind speed and wetness \cite{croft1995field}. In order to first rank these features \wrt their impact on the berry temperature, we train a random forest classifier \cite{liaw2002classification} and configure it to return feature importance scores. 
From the collected dataset and results presented in Section~\ref{sec:results}, we observe that two of the top-3 features are solar irradiance and humidity.
This shows the importance of predicting irradiance, since it varies frequently due to cloud movement, unlike the relatively constant humidity feature. 
Assuming constant humidity in the near-term (10-30 min), we can then predict future berry temperature with reasonable accuracy from the predicted future irradiance values. 
Finally, from this study, we complete our framework with a random forest \cite{liaw2002classification} and a multilayer perceptron (MLP) \cite{rosenblatt1961principles,ruck1990feature} to predict berry temperature from selected features.




\vspace{-0.4em}
\section{Results}
\label{sec:results}
\vspace{-0.4em}

\subsection{Cranberry Counting and Segmentation Results}
\begin{figure*}[t!]
\centering
\begin{tabular}{c@{\hspace{0.5em}}c@{\hspace{0.5em}}c}
    \includegraphics[width=0.28\linewidth]{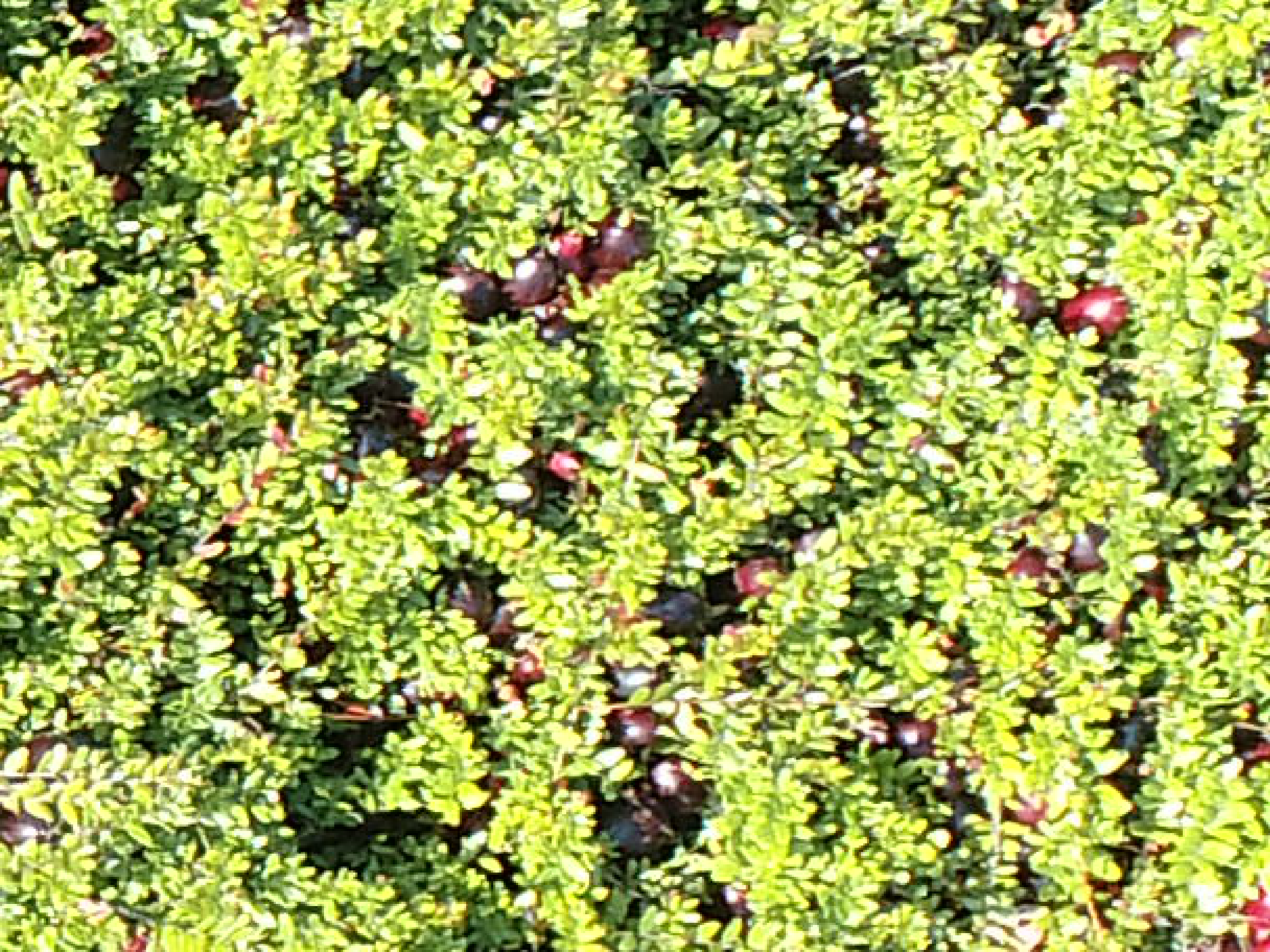} &   
    \includegraphics[width=0.28\linewidth]{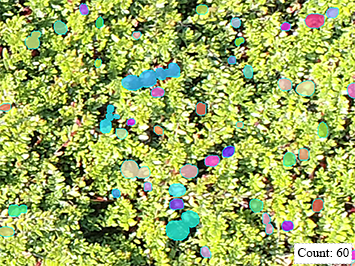} &   
    \includegraphics[width=0.28\linewidth]{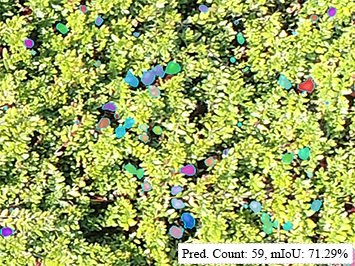} \\
    \includegraphics[width=0.28\linewidth]{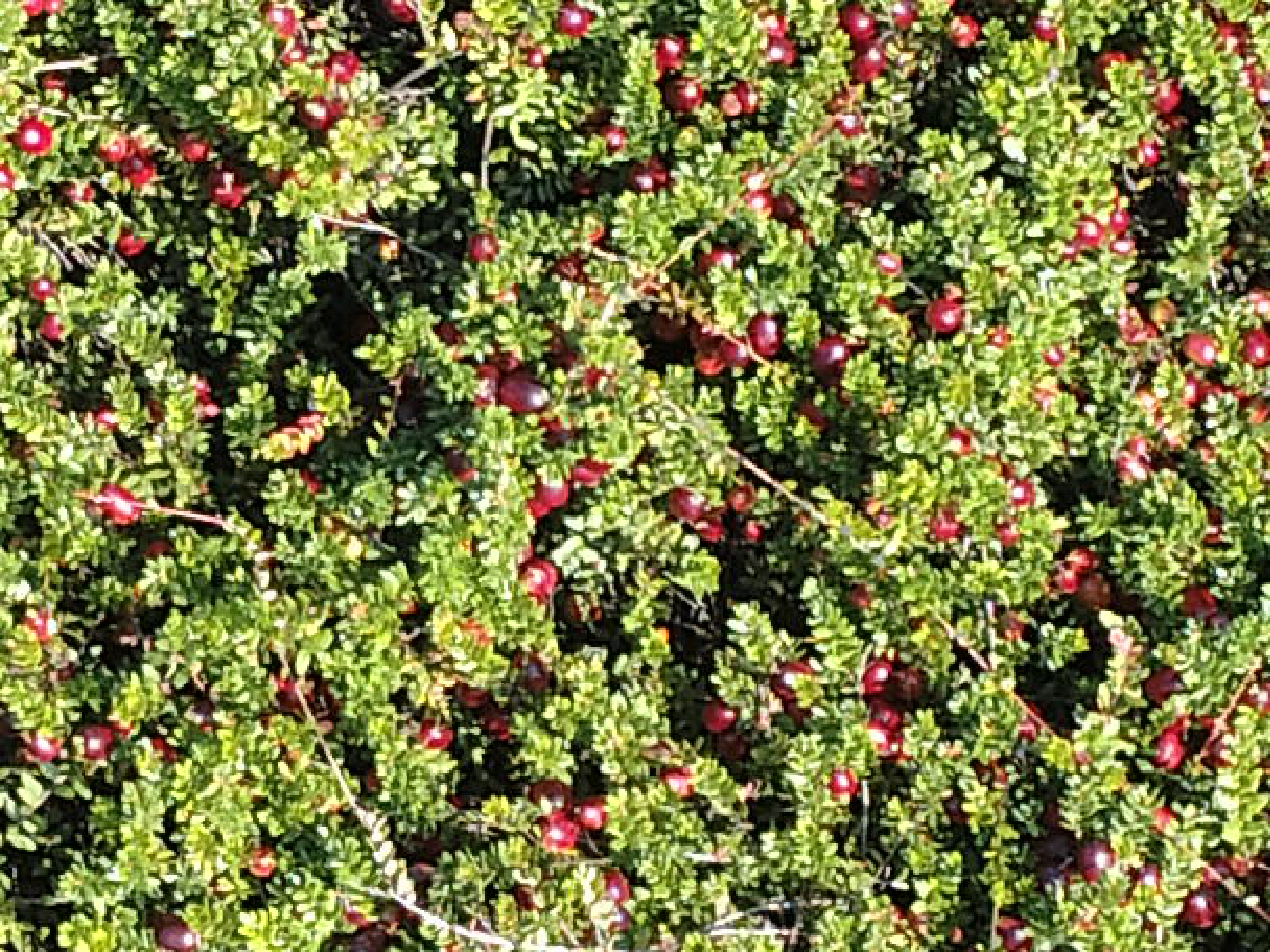} & 
    \includegraphics[width=0.28\linewidth]{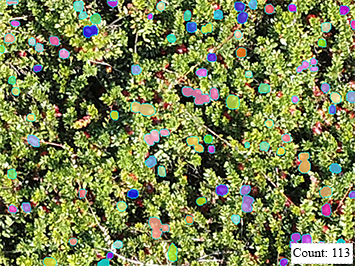} &
    \includegraphics[width=0.28\linewidth]{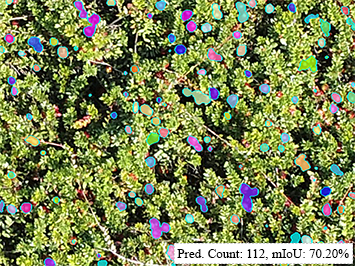} \\
    \includegraphics[width=0.28\linewidth]{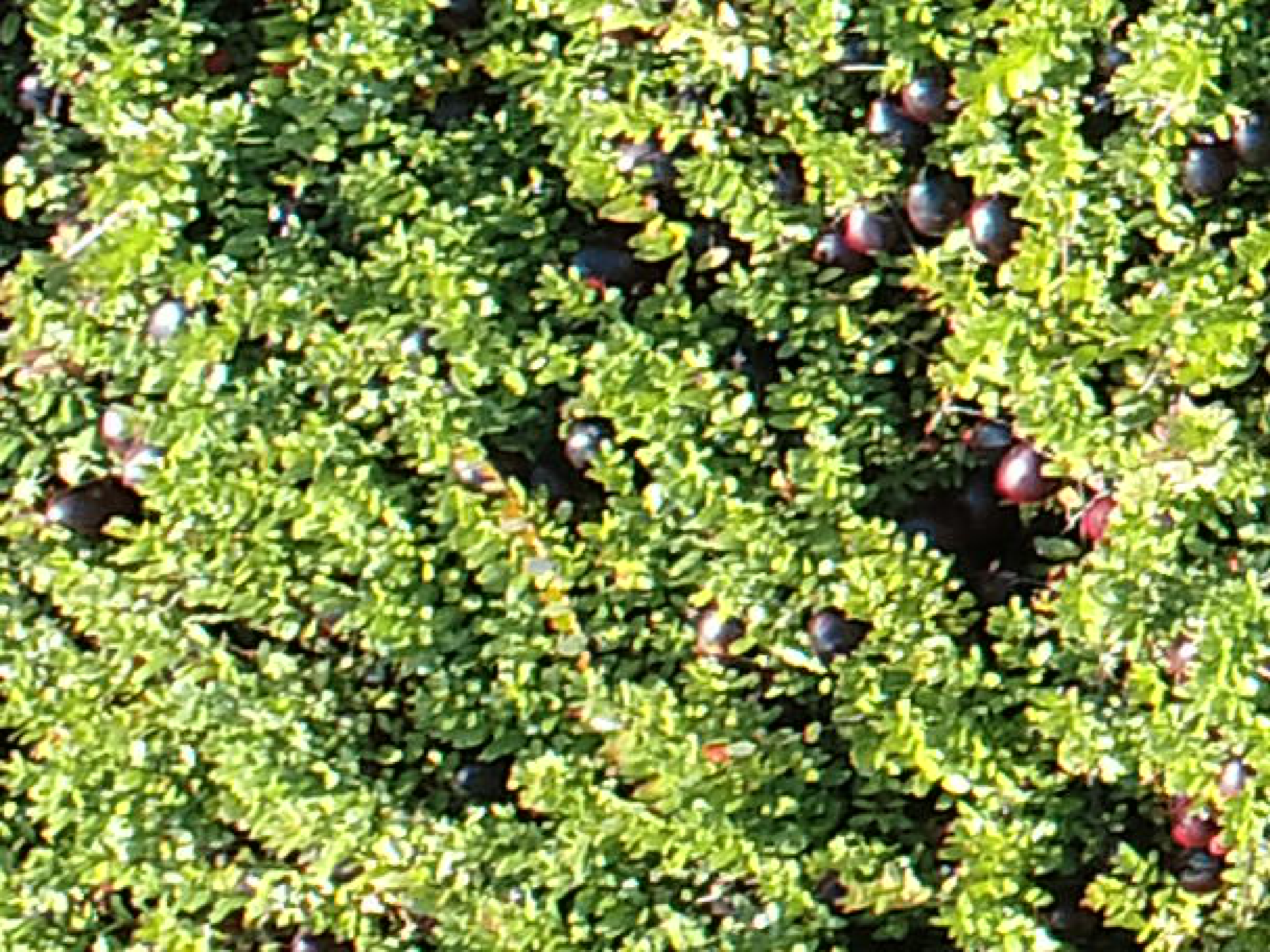} & 
    \includegraphics[width=0.28\linewidth]{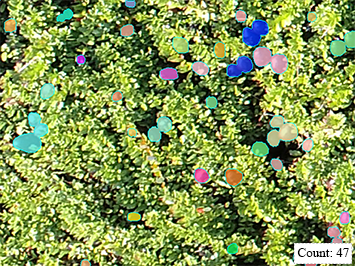} &
    \includegraphics[width=0.28\linewidth]{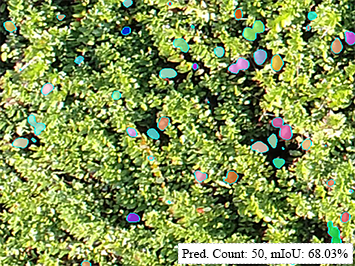} \\
    Input Image & Ground Truth  & Triple-S Network
\end{tabular}
\caption{Results of cranberry counting and segmentation using Triple-S method which is integrated in this application. Best viewed in color and zoomed.}
\vspace{-0.4cm}
\label{fig:cranberry_qual_results}
\end{figure*}

Qualitative and quantitative results are shown in Figure~\ref{fig:cranberry_qual_results} and Table~\ref{table:cranberry_table} respectively for the cranberry counting and segmentation module used in this application. Quantitative results are evaluated using mean intersection over union (mIoU) for segmentation performance, and mean absolute error (MAE) for counting performance. We use the Cranberry Aerial Imagery Dataset (CRAID) \cite{akiva2020finding} for training, validation and testing. The dataset provides 1022 images with center point annotations for training, and 231 images with pixel-wise annotation for evaluation. It can be seen that cranberries are correctly localized and separated into instances for accurate counting. Table \ref{table:cranberry_table} reports the metrics over the testing dataset for this method. In practice, each count is tied to an image location. Using a sequence of images covering an entire cranberry bog, we can generate a count density map for that bog. This provides farmers with essential information pertaining to exposed, high risk regions containing many exposed cranberries.  

\subsection{Irradiance Prediction Results}

\paragraph{Cloud Data set.}
Rigorous evaluation of any segmentation (and motion estimation) algorithm requires annotated image data sets. Most research groups working on sky/cloud segmentation use their own proprietary images that are not publicly available. To date, only two publicly available annotated sky image data sets are available. The HYTA database \cite{li2011hybrid} consists of 32 distinct images and Singapore Whole Sky IMaging SEGmentation Data set (SWIMSEG) \cite{dev2016color} consists of 1,013 images of captured over a period of 22 months. %
However, both datasets are too small for training any deep learning algorithms and do not provide time-series data for motion estimation.


We collected our own dataset of hemispheric sky images ($180^\circ$/$360^\circ$) at an interval of 5 seconds spread over \textapprox12 hrs. per day with Mobotix MX-Q24 fish eye camera. Corresponding irradiance measurements were collected using Kipp \& Zonen SMP11 pyranometer in unit of $watt/m^2$. The camera is calibrated using the OcamCalib Toolbox \cite{scaramuzza2006toolbox} with 11 checkerboard images. The results of the calibration are camera intrinsic matrix and ﬁsheye camera model that are used for projecting the original fisheye camera image in spherical coordinate to Cartesian coordinate for further image processing. For generating the segmentation label map, we apply standard color-based segmentation algorithm \cite{changShortTermCloud2019} which creates a noisy label data for training deep learning algorithms. 

\vspace{-1.2em}
\paragraph{Results.}

We tested our framework on a data set featuring different time of the day and month of the year. The training set was composed of image sequences and their noisy labels (\textapprox10,590 pairs) captured over the months of June and September; whereas the test set was made of sequences from May and August. Random image augmentations were performed to virtually increase the variability of the training data (random cropping, image flipping, brightness and contrast jitters) \cite{planche2019hands}. The network was trained until convergence using Adam optimizer with a learning rate of 0.001. 

\begin{table}[t!]
\centering
\resizebox{0.5\textwidth}{!}{%
    \begin{tabular}{ccc}
     \toprule
     \textbf{Method}  & \textbf{MAE}$^\downarrow$ & \textbf{mIoU (\%)}$^\uparrow$\\
      \midrule
      U-Net \cite{ronneberger2015u} & 18.67 & 60.61\\
      LC-FCN \cite{laradji2018blobs} & 17.46 & 61.97\\
      Triple-S Network ($\mathcal{L}_{seg} + \mathcal{L}_{split} + \mathcal{L}_{count}$) & \textbf{13.46} & \textbf{62.54}\\
     \bottomrule\\
\end{tabular}
}
\vspace{-1.0em}
\caption{Mean Intersection over Union (mIoU)
and Mean Absolute Error (MAE) 
accuracies on Cranberry Aerial Imagery Dataset 
\cite{akiva2020finding}
(values:  ``$\downarrow$" = lower is better; ``$\uparrow$" = higher is better).
}
\vspace{-1.5em}
\label{table:cranberry_table}
\end{table}

\begin{figure*}
    \centering
    \includegraphics[width=0.95\textwidth]{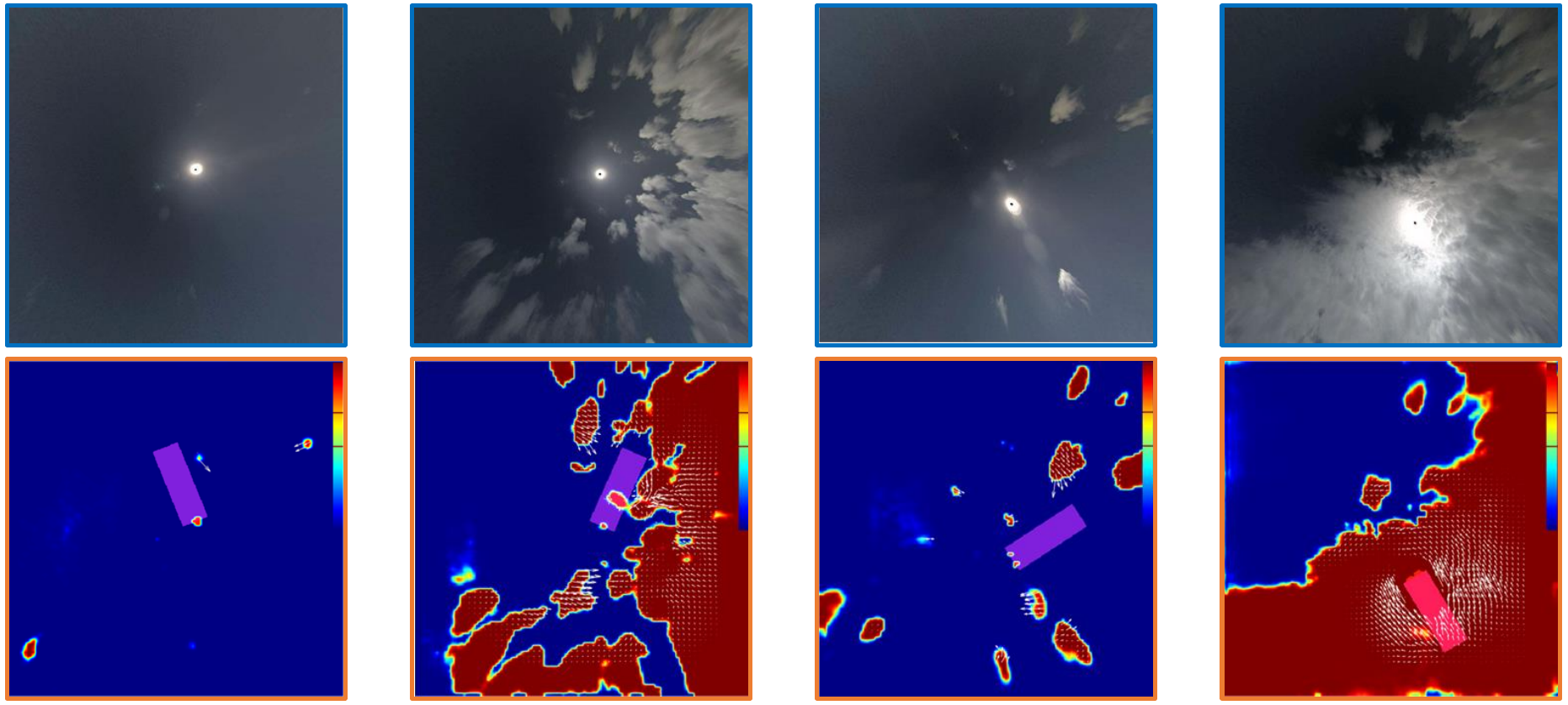}
    \caption{Results of cloud segmentation and motion estimation. Top row: original image; 
    bottom row: output of our framework. The color gradient represents the predicted cloud probability, the arrows the predicted cloud motion, and the purple rectangle the region of interest for 20-min irradiance forecast based on sun position and cloud motion.}
    \vspace{-1.5em}
    \label{fig:seg_result2}
\end{figure*}

\begin{figure}
    \centering
    \includegraphics[width=0.49\textwidth]{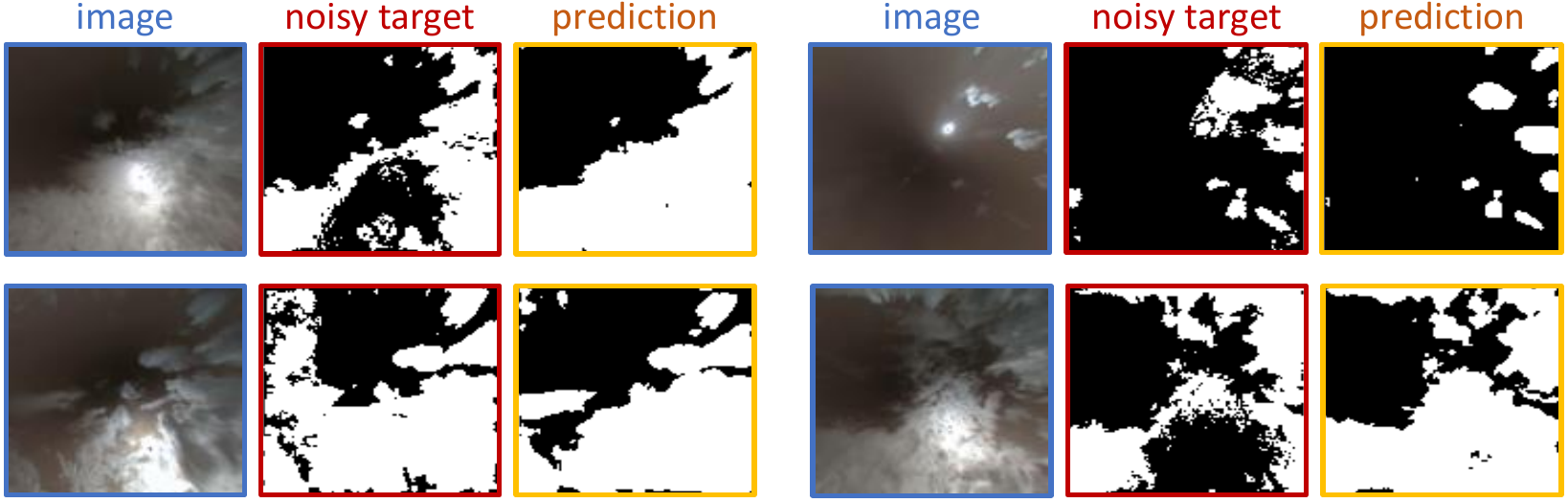}
    \caption{Segmentation results on test image sequences.}
    \vspace{-1.2em}
    \label{fig:seg_result1}
\end{figure}

Figure \ref{fig:seg_result1} shows qualitative results on the semantic cloud segmentation task. It can be observed how, despite being trained on noisy target labels, the pipeline achieves more consistent results, thanks to the additional supervisions (through motion estimation and sequence consistency). Figure \ref{fig:seg_result2} shows additional results of joint segmentation and motion estimation.

\vspace{-1em}

\begin{figure}
    \centering
    \includegraphics[width=0.45\textwidth]{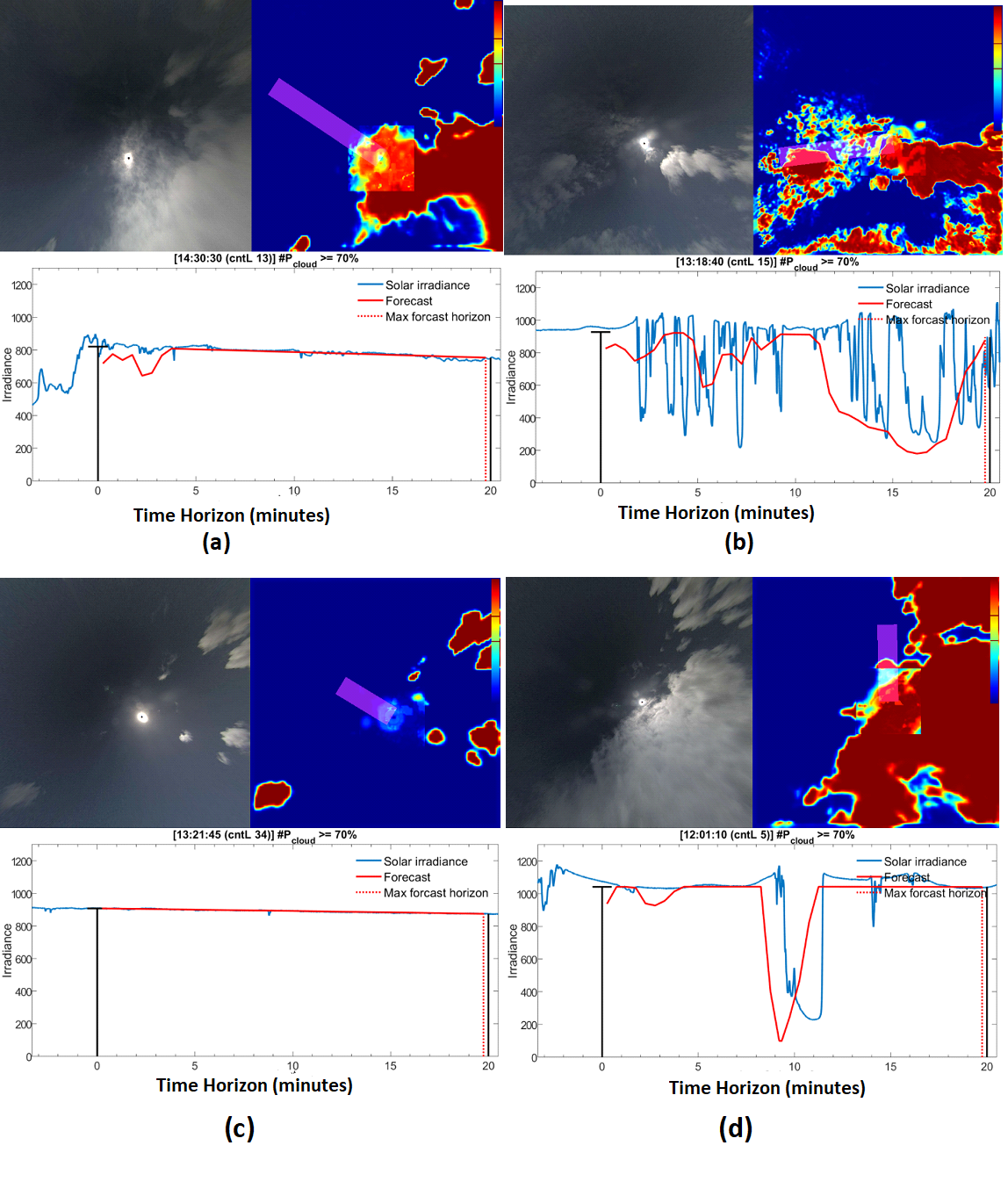}
    \vspace{-1.em}
    \caption{Irradiation prediction results in different cloud scenarios. Proposed framework could correctly match ground truth irradiation during (a) sun occlusion by thin clouds and (c) clear sky without any cloud coverage of the sun. (b) When sun occlusion happens by presence of frequent irregular clouds, our approach could closely predict the overall ground truth irradiation trend. However, minor variation in irradiation is not predicted. (d) Sudden pop-up cloud appearance was identified by the present method with limited time-difference.}
    \vspace{-1.5em}
    \label{fig:ir_result_combined}
\end{figure}

Due to the lack of exact ground-truth labels for the cloud segmentation task, meaningful quantitative results can only be obtained through indirect metrics. Synchronized pyrometer data captured with the sky data was used as ground-truth irradiance to evaluate the cloud segmentation-based irradiance prediction method. 
Three different metrics were used to evaluate the accuracy of the prediction over a time horizon of 20 minutes, namely MAPE (Mean Absolute Percentage Error), R-squared ($R^2$), and Fréchet distance. The ground truth and predicted series, resp.  $\widetilde{Y} = \{\widetilde{y_i}\}_{i=t}^{t+\tau}$ and $Y = \{y_i\}_{i=t}^{t+\tau}$, are normalized before comparison based on the maximum and minimum values of the series over the time period. 
The MAPE is defined as 
$\MAPE = \frac{1}{\tau+1}\sum_{i=t}^{t+\tau}{|\frac{\widetilde{y_i} - y_i}{\widetilde{y_i}}|}$. R-squared (explained variance) is defined as $R^2 = 1 - \frac{\|\widetilde{Y} - Y\|_2}{\|\widetilde{Y} - \overline{y}\|_2}$ with $\overline{y} = \frac{1}{\tau+1}\sum_{i=t}^{t+\tau}{\widetilde{y_i}}$ average of ground-truth series. The Fréchet distance is a measure of similarity between curves that takes into account the location and ordering of the points along the curves. It is defined as the minimum cord-length sufficient to join a point traveling forward along predicted curve and one traveling forward along ground truth curve, although the rate of travel for either point may not necessarily be uniform. If the Fréchet distance between two curves is small, it indicates that the curves are similar. 
Table \ref{tab:irradiance} demonstrates performance of the proposed system using these three metrics in several scenarios, with an accuracy increase attributable to our contributions \wrt network self-supervision.  
Prediction results on sampled sequences of different sky conditions are also shown in Figure~\ref{fig:ir_result_combined}. We can observe how cloud motion prediction (even if locally challenging due to lack of texture and complex dynamics) effectively guides the irradiance forecasting in varied sun occlusion scenarios. By extracting a region of interest within each sky image based on the sun position and overall cloud motion, only detected clouds that may occlude the sun in the time horizon are considered by the irradiance prediction module as mentioned in Section 3.3.

\begin{figure}
    \centering
    \includegraphics[width=0.5\textwidth]{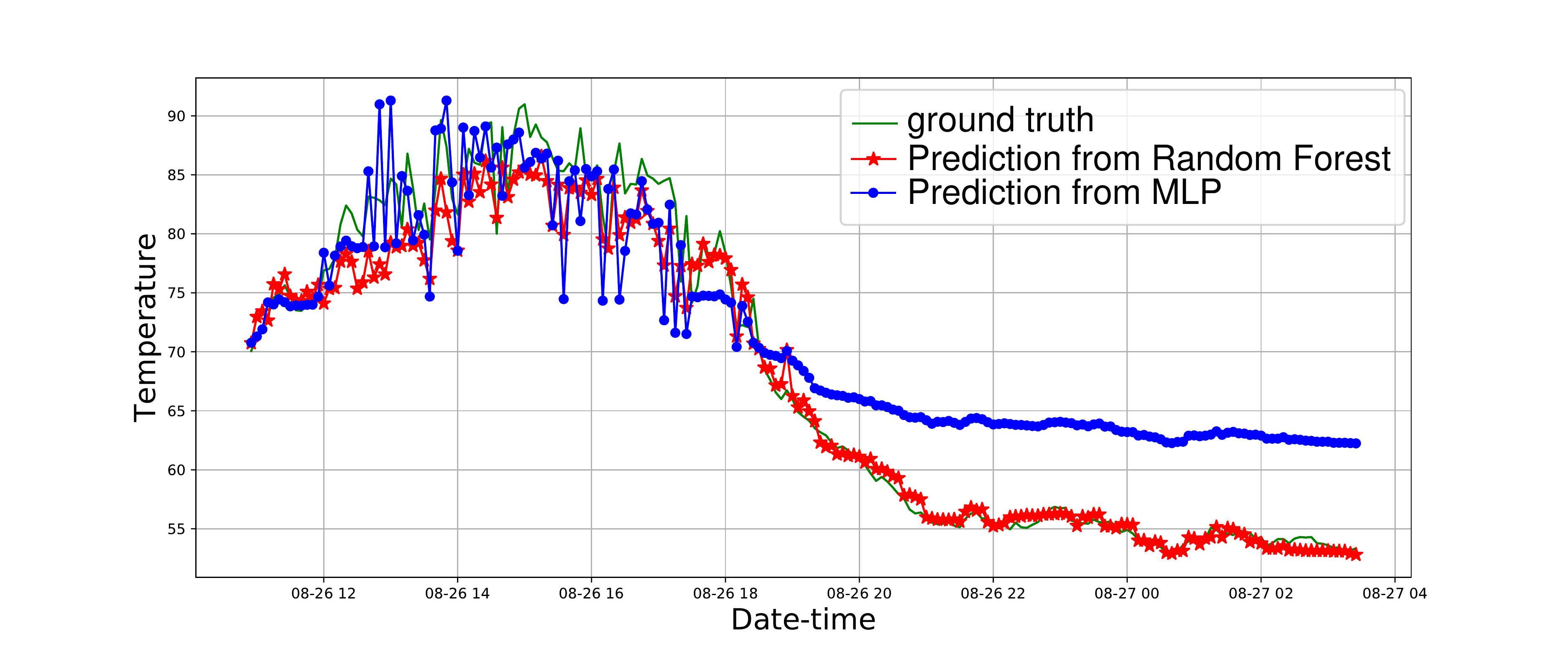}
    \caption{Berry temperature prediction results using Random Forest and MLP.}
    \label{fig:comp_model}
    \vspace{-1.5em}
\end{figure}


\begin{table*}[t]
\centering
\resizebox{1\linewidth}{!}{
\footnotesize
\def\arraystretch{1}
\begin{tabu}{@{}cccc|ccc|ccc|ccc@{}}
\toprule
\multicolumn{4}{c|}{\textbf{Components of Pipeline for Sky Image Analysis}} & \multicolumn{3}{c|}{\textbf{Irr. Forecast -- 5min}} & \multicolumn{3}{c|}{\textbf{Irr. Forecast -- 10min}} & \multicolumn{3}{c}{\textbf{Irr. Forecast -- 20min}} \\
\cmidrule(lr){1-4}\cmidrule(lr){5-7}\cmidrule(lr){8-10}\cmidrule(lr){11-13}
baseline \cite{qinJointLearningMotion2018a} & $\mathcal{L}_x^{w_{rev}}$ & $v \odot m$ & flow ref. \cite{lucas1981iterative} & MAPE $^\downarrow$ & $R^2$ $^\uparrow$ & Fréchet $^\downarrow$ & MAPE $^\downarrow$ & $R^2$ $^\uparrow$ & Fréchet $^\downarrow$ & MAPE $^\downarrow$ & $R^2$ $^\uparrow$ & Fréchet $^\downarrow$\\

\midrule
$\checkmark$ & & & & 10.01 & 0.77 & 0.501 & 16.92 & 0.79 & 0.634 & 29.18 & 0.67 & 0.765
\\

$\checkmark$ & $\checkmark$ & & & 9.13 & 0.75 & \textbf{0.477} & 14.53 & 0.72 & 0.593 & 21.23 & 0.79 & 0.664
\\

$\checkmark$ & $\checkmark$ & & $\checkmark$ & 8.86 & 0.92 & 0.479 & 12.89 & 0.76 & 0.602 & \textbf{20.36} & 0.81 & \textbf{0.564}
\\

$\checkmark$ & & $\checkmark$ & & 8.88 & 0.93 & 0.485 & 13.76 & \textbf{0.84} & 0.596 & 22.52 & 0.84 & 0.655
\\

$\checkmark$ & & $\checkmark$ & $\checkmark$ & \textbf{8.41} & \textbf{0.95} & 0.507 & \textbf{12.32} & \textbf{0.84} & \textbf{0.573} & 22.84 & \textbf{0.86} & 0.592
\\

\bottomrule
\end{tabu}
}
\vspace{0.15em}  
\caption{Ablation study over irradiance prediction results. We compare the accuracy on the end task of irradiance forecasting for the baseline \cite{qinJointLearningMotion2018a} and our extensions, \ie, the  backward/forward consistency loss $\mathcal{L}_x^{w_{rev}}$ and the masking of motion maps $v$ with cloud probability maps $m$. We also consider the refinement of motion maps using Lucas-Kanade method \cite{lucas1981iterative} (values:  ``$\downarrow$" = lower is better; ``$\uparrow$" = higher is better).}
\label{tab:irradiance} 
\vspace{-1em}   
\end{table*}

\subsection{Berry Temperature Prediction Results}

We synchronously collected berry temperature and weather data over a period of 2.5 months with 5-minute sampling rate. 
Collected weather data contains information about ambient temperature (\textdegree F), wind speed (mile/hr), gust speed  (mile/hr), wind direction (0-360\textdegree), relative humidity, dew point (\textdegree F), rain (in), wetness (\%), solar irradiance ($w/m^2$). Our current temperature prediction uses ground truth solar irradiance (measured by a pyranometer) as input, since our sky image dataset was collected from a separate location.  Ground truth internal temperature data for training is obtained with a synthetic cranberry temperature probe as shown in Figure~\ref{fig:teaser}. 
Training and testing samples for temperature prediction were split 70-30 over temporally-disjoint periods, performing 5-fold cross validation.
Predicted berry temperature values against ground truth using the two models are shown in Figure \ref{fig:comp_model}. Comparative analysis of the two machine learning models, namely random forest and MLP, using three metrics (MAE, MAPE, and $R^2$ defined in previous sections) is shown in Table~\ref{tab:berry_temp}. It can be observed from the table as well as from the figure, that the random forest based model prediction is better than the MLP model.
The features selected by the random forest method in order of importance are: solar irradiance, relative humidity, dew point, gust speed, wind speed, wind direction and wetness. 


\begin{table}[t]
\centering
\begin{tabular}{p{2.7cm}|cx{1.3cm}cx{1.3cm}cx{1.3cm}} 
\toprule
\textbf{Method} & \textbf{MAPE} $^\downarrow$ & \textbf{MAE} $^\downarrow$  & $\mathbf{R^2}$ $^\uparrow$ \\
\midrule
Random Forest & 0.02 &  1 & 0.98 \\
MLP & 0.04 & 3 & 0.91 \\ 
\bottomrule
\end{tabular} 
\vspace{0.5em}   
\caption{Method comparison for berry temperature prediction.}
\label{tab:berry_temp} 
\vspace{-1.5em}  
\end{table}



\vspace{-0.5em}
\section{Discussion and Conclusions}

The proposed  crop-monitoring technology  for small field farming  achieves  advantages of low cost, high efficiency and precision.
By quantifying crop exposure and risk in near real-time, the proposed framework provides  growers the ability to optimize resources and
sustain the economic viability of the farm.

There are several major challenges. Variations in cloud appearance, cloud motion, and abrupt deformation makes cloud segmentation a challenging task. The presence of anomalies covering the camera field of view such as trees, birds, and debris,  can interfere with cloud coverage prediction. 
For temperature prediction, training data from the synthetic cranberry temperature probe is representative of real cranberry temperature but not exact due to the probe position located at a different height from the imaged cranberries. 
Cranberry segmentation and counting are affected by change in berry appearance (shape, size, and albedo) and partial occlusions. 
To expand the technology into new fruit application areas in the future, we need to handle technological issues related to different fruit properties and acquisition of the corresponding large-scale datasets to make the system generalizable.
While challenges remain,
the  proposed technology is a significant step towards real-time robust automatic fruit health monitoring, thus promoting the development of agricultural automation equipment and precision agriculture systems.

\vspace{-1.em}

{\small
\paragraph{Acknowledgements} This project was sponsored by the USDA NIFA AFRI Award Number: 2019-67022-29922.
}

{\small
\bibliographystyle{ieee_fullname}
\bibliography{egbib}
}

\end{document}